%% file: main.tex
\documentclass[10pt,twocolumn,letterpaper]{article}
\usepackage[accsupp]{axessibility} % Improves PDF readability for those with disabilities.
\usepackage{iccv}
\usepackage{times}
\usepackage{epsfig}
\usepackage{graphicx}
\usepackage{amsmath}
\usepackage{amsfonts}
\usepackage{ifthen} 
\usepackage{booktabs}
\usepackage{amssymb}
\usepackage{mmstyle}
\usepackage{mathtools} % The \coloneq they refer to comes from here, yes?
\usepackage{breqn}
\usepackage{caption}
\usepackage{enumitem}% In any case, if you remove the mathtools package,
\usepackage[pagebackref=true,breaklinks=true,letterpaper=true,colorlinks,bookmarks=false]{hyperref}
\usepackage{multirow}

\iccvfinalcopy % *** Uncomment this line for the final submission

\def\iccvPaperID{1680} % *** Enter the ICCV Paper ID here

\input{utils/definitions.tex}
\input{utils/math_commands.tex}

\begin{document}

%%%%%%%%% TITLE
\title{\METHODNAME: Deformable Neural Radiance Fields for 3D Toonification}

\author{
Junzhe Zhang$^{1,3}$\footnotemark[1] \quad
Yushi Lan$^{1}$\footnotemark[1]  \quad
Shuai Yang$^{1}$  \quad
Fangzhou Hong$^{1}$ \\
Quan Wang$^{3}$  \quad
Chai Kiat Yeo$^{2}$ \quad
Ziwei Liu$^{1}$ \quad
Chen Change Loy$^{1}$ \\
$^{1}$S-Lab, Nanyang Technological University  \hspace{4pt} \\
$^{2}$Nanyang Technological University  \hspace{4pt}  $^{3}$SenseTime Research \\
{\tt\small \{shuai.yang,asckyeo,ziwei.liu,ccloy\}@ntu.edu.sg} \\
{\tt\small \{junzhe001,yushi001, fangzhou001\}@e.ntu.edu.sg} \hspace{4pt}
{\tt\small \{wangquan\}@sensetime.com}
}

\twocolumn[{%
\renewcommand\twocolumn[1][]{#1}%
\maketitle
\vspace{-12mm}
\begin{center}
\centering
\includegraphics[width=1\linewidth]{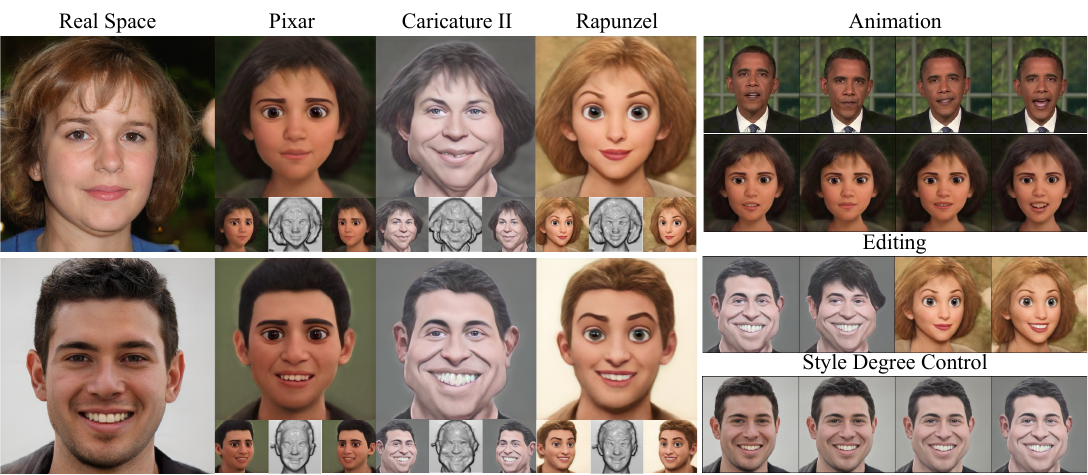}
\vspace{-5mm}
\captionof{figure}{We propose \METHODNAME, an efficient 3D toonification framework which supports geometry-texture decomposed toonification over multiple styles.
\METHODNAME{} is fine-tuning free and could be easily extended to a series of downstream applications designed for the pre-trained GAN, \eg, animation given a driving video, semantic attributes editing ($+$ bangs and smile), and flexible style degree control.
}
\vspace{-1mm}
\label{fig:teaser}
\end{center}%
}]

\renewcommand{\thefootnote}{\fnsymbol{footnote}}
\footnotetext[1]{Equal contribution.}

\maketitle
\ificcvfinal\thispagestyle{empty}\fi

% \input{iccv2023AuthorKit/tabs_and_figs/fig_teaser}

%%%%%%%%% BODY TEXT
%\input{iccv2023AuthorKit/tabs_and_figs/fig_teaser}
\input{sections/0_abstract}
\input{sections/1_introduction}

\input{sections/2_related_work}
\input{sections/3_method}
\input{sections/4_experiment}
\input{sections/5_conclusion}

\noindent
\textbf{Acknowledgement.} This study is supported under the RIE2020 Industry Alignment Fund Industry Collaboration Projects (IAF-ICP) Funding Initiative, as well as cash and in-kind contribution from the industry partner(s). It is also supported by Singapore MOE AcRF Tier 2 (MOE-T2EP20221-0011, MOE-T2EP20221-0012) and NTU NAP Grant.

\appendix
\renewcommand\thesection{\Alph{section}}
\input{sections/6_appendix}

\clearpage

{\small
\bibliographystyle{ieee_fullname}
\bibliography{egbib}
}

% \clearpage
% \input{sections/6_appendix}

\end{document}

%% file: utils/definitions.tex
\renewcommand{\vec}[1]{\boldsymbol{#1}}
\newcommand{\mat}[1]{\mathbf{#1}}

\newcommand{\loss}{\mathcal{L}}
\newcommand{\lossweight}{\lambda}

\newcommand{\nerf}{\mathcal{N}}

\newcommand{\point}{\vec{x}}

\newcommand{\code}{\vec{z}}

\newcommand{\dist}{d}

\def\METHODNAME{\textsc{DeformToon3D}}

\newcommand{\heading}[1]{\noindent\textbf{#1.}}

\newlength\savewidth
  
\newcommand{\rendererflag}{\mathbf{0}}
\newcommand{\decoderflag}{\mathbf{1}}

\newcommand{\generator}{G}
\newcommand{\rendererG}{\generator_{\rendererflag}}
\newcommand{\decoderG}{\generator_{\decoderflag}}

\newcommand{\wspace}{\mathcal{W}}

\newcommand{\mmspace}{\mathcal{M}}

\newcommand{\view}{\bm{v}}

\newcommand{\campose}{\bm\xi}

\newcommand{\datasetsample}{\mathcal{X}}

\newcommand{\image}{\mat{I}}

\newcommand{\imagePredicted}{\hat{\image}}

\newcommand{\zz}{{\bf z}}
\newcommand{\ww}{{\bf w}}
\newcommand{\mm}{{\bf m}}

\newcommand{\zcode}{\zz}
\newcommand{\wcode}{\ww}

\newcommand{\RN}[1]{%
  \textup{\uppercase\expandafter{\romannumeral#1}}%
}

%% file: utils/math_commands.tex
\usepackage{amsmath,amsfonts,bm}

\def\eqref#1{equation~\ref{#1}}
\def\1{\bm{1}}

\def\vb{{\bm{b}}}

\def\vf{{\bm{f}}}

\def\mW{{\bm{W}}}

\DeclareMathAlphabet{\mathsfit}{\encodingdefault}{\sfdefault}{m}{sl}
\SetMathAlphabet{\mathsfit}{bold}{\encodingdefault}{\sfdefault}{bx}{n}

\DeclarePairedDelimiter\abs{\lvert}{\rvert}%
\DeclarePairedDelimiter\norm{\lVert}{\rVert}%

%% file: sections/0_abstract.tex
% !TEX root = ../main.tex
%%%%%%%%% ABSTRACT
\begin{abstract}

In this paper, we address the challenging problem of 3D toonification, which involves transferring the style of an artistic domain onto a target 3D face with stylized geometry and texture. 
Although fine-tuning a pre-trained 3D GAN on the artistic domain can produce reasonable performance, this strategy has limitations in the 3D domain. In particular, fine-tuning can deteriorate the original GAN latent space, which affects subsequent semantic editing, and requires independent optimization and storage for each new style, limiting flexibility and efficient deployment. 
To overcome these challenges, we propose \METHODNAME, an effective toonification framework tailored for hierarchical 3D GAN. Our approach decomposes 3D toonification into subproblems of geometry and texture stylization to better preserve the original latent space. 
Specifically, we devise a novel StyleField that predicts conditional 3D deformation to align a real-space NeRF to the style space for geometry stylization. 
Thanks to the StyleField formulation, which already handles geometry stylization well, texture stylization can be achieved conveniently via adaptive style mixing that injects information of the artistic domain into the decoder of the pre-trained 3D GAN. Due to the unique design, our method enables flexible style degree control and shape-texture-specific style swap. 
Furthermore, we achieve efficient training without any real-world 2D-3D training pairs but proxy samples synthesized from off-the-shelf 2D toonification models.
Code is released at \textcolor{magenta}{\url{https://github.com/junzhezhang/DeformToon3D}}.

\end{abstract}

%% file: sections/1_introduction.tex
% !TEX root = ../main.tex
% \vspace{-5mm}
\section{Introduction}
\label{sec:intro}

Artistic portraits are prevalent in various applications such as comics, animation, virtual reality, and augmented reality. 
In this work, our main objective is to propose an effective approach for 3D-aware artistic toonification, a critical problem that involves transferring the style of an artistic domain onto a target 3D face with stylized geometry and texture. 
The task opens up potential applications for quick high-quality 3D avatar creation based on a photograph with the style of a designated artwork, which would typically require highly professional handcraft skills.

Substantial progress has been made in automatic portrait style transfer over 2D images. Starting with image style transfer~\cite{gatys2016image,selim2016painting,liao2017visual,kim2020deformable} and image-to-image translation~\cite{kim2019u,li2021anigan,chong2021gans,cao2018carigans,shi2019warpgan}, recent advancements in StyleGAN-based generators~\cite{karras2019style,karras2020analyzing} have shown their potential in high-quality toonification via efficient transfer learning~\cite{pinkney2020resolution}. Specifically, a pre-trained StyleGAN generator on face images is fine-tuned to transfer to the artistic portrait domain. With the progress of 3D-aware GANs~\cite{Chan2021EG3D,orel2021stylesdf}, researchers have extended this pipeline to 3D with well-designed domain adaptation frameworks~\cite{Jin2022Dr3DA3,lan2022e3dge,kim2022datid3d,Abdal20233DAvatarGANBD}, enabling remarkable 3D portrait toonification.

% problem of the previous methods and our motivation
Although fine-tuning the pre-trained StyleGAN-based model for toonification achieves superior quality, it has several limitations.
\textbf{First}, fine-tuning the pre-trained generator shifts its generative space from the real face domain to the artistic portrait domain at the cost of deteriorating the original GAN latent space. With tremendous off-the-shelf tools~\cite{Shen2020InterFaceGANIT,tewari2020stylerig} trained for the original GAN space, altering the well-learned style space would affect the performance of downstream applications over the toonified portrait, \eg, semantic editing.
\textbf{Second}, despite fine-tuning-based domain adaptations have been thoroughly investigated for 2D GANs~\cite{Pinkney2020ResolutionDG,wu2021stylealign}, applying this technique to 3D GANs fails to leverage the full potential of the architecture of 3D generator models~\cite{Chan2021EG3D,orel2021stylesdf} for characterizing view-consistent shape and high-frequency textures in the artistic domain.
\textbf{Third}, it is inevitable to fine-tune a heavy generator for each new style, which requires hours of training time and additional storage. This limitation affects scalability when deploying dozens of fine-tuned generators for real-time user interactions.
%It also limits the application of mixing the geometry toonification of one style and texture toonification of another.
Therefore, 3D toonification remains a challenging task that requires further exploration.

To better preserve the pre-trained GAN latent space and to better exploit the 3D GAN generator, we propose \textbf{\METHODNAME}~that decomposes geometry and texture stylization into more manageable subproblems.
In particular, unlike conventional 3D toonification approaches that fine-tune the whole 3D GAN generator following existing 2D fine-tuning schemes, we carefully consider the characteristics of 3D GANs to decompose the stylization of geometry and texture domains.
To achieve geometry stylization, we introduce a novel \textbf{StyleField} on top of a pre-trained 3D generator to deform each point in the style space to the pre-trained real space guided by an instance code. This allows for easy extension to multiple styles with a single stylization field by introducing a style code to guide the deformation.
Since StyleField already handles geometry stylization well, texture stylization can be easily achieved through adaptive style mixing which injects artistic domain information into the network for effective texture toonification. 
Notably, our unique design enables training of the method at minimal cost using synthetic paired data with realistic faces generated by a pre-trained 3D GAN and corresponding paired stylized data generated by an off-the-shelf 2D toonification model~\cite{yang2022Pastiche}.
%Supervisions on both texture space and 3D space are imposed for high-quality 3D toonification.

The proposed \METHODNAME~achieves high-quality geometry and texture toonification over a vast variety of styles, as demonstrated in Fig.\ref{fig:teaser}. Additionally, our approach preserves the original GAN latent space, enabling compatibility with existing tools built on the real face space GAN, including inversion~\cite{lan2022e3dge}, editing~\cite{Shen2020InterFaceGANIT}, and animation~\cite{tewari2020stylerig}. Furthermore, our design significantly reduces the storage footprint by requiring only a small stylization field with a set of AdaIN parameters for artistic domain stylization. In summary, our work makes the following contributions:
\begin{itemize}
\item We propose a novel StyleField that separates geometry toonification from texture, providing a more efficient method for modeling 3D shapes than fine-tuning and enabling flexible style control.
\item We present an approach to achieve multi-style toonification with a single model, facilitating cross-style manipulation and reducing storage footprint.
\item We introduce a full synthetic data-driven training pipeline that offers an efficient and cost-effective solution to training the model without requiring real-world 2D-3D training pairs.
\end{itemize}

%% file: sections/2_related_work.tex
% !TEX root = ../main.tex
% \vspace{-2mm}
\section{Related Work}
\noindent
\textbf{3D Generative Models.}
% ==== 2D works
Inspired by the success of Generative Adversarial Networks (GAN)\cite{Goodfellow2014GenerativeAN} in generating photorealistic images\cite{karras2019style,Brock2019LargeSG,karras_analyzing_2020}, researchers have been making efforts towards 3D-aware generation~\cite{NguyenPhuoc2019HoloGANUL,platogan,pan_2d_2020}.
Starting with explicit intermediate shape representations, such as voxels~\cite{NguyenPhuoc2019HoloGANUL,platogan} and meshes~\cite{pan_2d_2020}, which lack photorealism and are memory-inefficient, researchers have recently shifted towards using implicit functions~\cite{park_deepsdf_2019,Mescheder2019OccupancyNetwork,chen_learning_2019} along with physical rendering processes~\cite{sitzmann_scene_2019,mildenhall2020nerf} as intrinsic 3D inductive biases.
Among these approaches, 3D generative models~\cite{Chan2021piGANPI,Schwarz2020NEURIPS} extended from neural radiance fields (NeRF)~\cite{mildenhall2020nerf} have demonstrated impressive view-consistency in synthesized results. 
While the original NeRF is limited to modeling static scenes, recent research has introduced deformation fields to enable NeRF to model dynamic volumes~\cite{park2021nerfies,park2021hypernerf,Tewari2022Disentangled3DLA,lan2022ddf}.
To increase the resolution of generated images, recent studies~\cite{Chan2021EG3D,hong2023eva3d} have resorted to voxel-based representations or adopted a hybrid design~\cite{GIRAFFE,orel2021stylesdf,Chan2021EG3D,gu2021stylenerf}. 
This hybrid design involves a cascade model $\generator=\decoderG\circ\rendererG$ pairing a 3D generator $\rendererG$ with a 2D super-resolution decoder $\decoderG$. 
Both $G_0$ and $G_1$ follow the style-based architecture~\cite{karras2019style,karras2020analyzing} to accept a latent code $\wcode$ to control the style of the generated object.
By super-resolving the intermediate low-resolution 2D features produced by the $\rendererG$ with the $\decoderG$, the hybrid design achieves view-consistent synthesis at high resolution, \eg, $1024^2$. 

\heading{Domain Adaptation for StyleGAN Toonification}
% with Domain Adaptation.
Researchers have typically employed domain adaptation in 2D space to achieve toonification with StyleGANs~\cite{karras2019style,karras_analyzing_2020}.
Typically, a portrait StyleGAN pre-trained on real images~\cite{CelebAMask-HQ,karras2018progressive} is fine-tuned on an artistic domain dataset to generate toonified faces.
Building upon this straightforward framework~\cite{Pinkney2020ResolutionDG}, a series of in-depth research has been conducted to further improve style control~\cite{yang2022Pastiche,yang2022vtoonify}, choices of latent code~\cite{Song2021AgileGANSP}, few-shot training~\cite{Ojha2021FewshotIG}, and text-guided adaptation~\cite{Gal2021StyleGANNADACD}.

Pre-trained 3D GANs offer high-quality generation and thus have the potential to facilitate downstream applications such as portrait stylization through 3D GAN inversion~\cite{lan2022e3dge,Chan2021EG3D}.
To extend 2D StyleGAN domain adaptation to 3D, CIPS-3D~\cite{zhou2021CIPS3d} proposed fine-tuning only the super-resolution decoder module for view-consistent texture toonification. However, it is limited to texture toonification since the 3D generator is left unchanged. 
Dr.3D~\cite{Jin2022Dr3DA3}, E3DGE~\cite{lan2022e3dge}, and 3DAvartarGAN~\cite{Abdal20233DAvatarGANBD} fine-tune the entire 3D generator~\cite{Chan2021EG3D,orel2021stylesdf} for both geometry and texture toonification, while DATID-3D~\cite{kim2022datid3d} leverages a Stable Diffusion~\cite{Rombach2021HighResolutionIS} generated corpus for text-guided domain adaptation.
While these methods yield impressive results, they come with limitations, as as they require costly fine-tuning and independent model storage for each new style.
Furthermore, previous fine-tuning-based toonification methods suffer from limited generality due to the incompatibility with abundant editing techniques developed for the original StyleGAN latent space~\cite{Shen2020InterFaceGANIT,shen2021closed,Patashnik2021StyleCLIPTM}.
In contrast, \METHODNAME~fully preserves the original 3D GAN latent space, which makes it intrinsically compatible with the editing methods trained for the original StyleGAN space. It achieves comparable visual quality while being $10$ times more storage-efficient than previous methods.

%% file: sections/3_method.tex
% !TEX root = ../main.tex
\input{tabs_and_figs/fig_framework}
\section{\METHODNAME}

We present the framework of \METHODNAME~in Fig.~\ref{fig:framework}.
Our approach begins with a typical hybrid 3D-aware design~\cite{orel2021stylesdf,Chan2021EG3D} that generates real-domain faces, and reformulates it to a 3D toonification framework.
The approach starts with a cascade model $\generator=\decoderG\circ\rendererG$, which pairs a 3D generator $\rendererG$ with a 2D super-resolution decoder $\decoderG$.
The generator $\rendererG$ captures the underlying geometry with the {instance code} $\wcode$ and camera pose $\campose$, and produces an intermediate feature map $\mathbf{F}$ with volume rendering~\cite{mildenhall2020nerf}. Then, $\decoderG$ upsamples $\mathbf{F}$ to obtain a high-resolution image $\image$ with high-frequency details added.
To adapt $G$ from the real domain to the artistic or cartoon domain, existing methods~\cite{Jin2022Dr3DA3,Abdal20233DAvatarGANBD,zhou2021CIPS3d,lan2022e3dge} view $\decoderG$ and $\rendererG$ as a whole and
simply fine-tune the pre-trained $G$, failing to take advantage of the decomposed characteristics of the hybrid framework design.
By comparison, \METHODNAME~fully exploits this cascaded synthesis process by using a novel StyleField module for geometry stylization, which in turn also benefits appearance stylization, allowing it to adopt a simple adaptive style mixing strategy.
In Sec.~\ref{sec:method:geo_toonification}, we elaborate the proposed StyleField along with the pre-trained $\rendererG$ to handle geometry stylization.
In Sec.~\ref{sec:method:tex_toonification}, we explain how the adaptive style mixing injects the style of the target domain into $\decoderG$ to achieve texture stylization.
Lastly, we present our training pipeline in Section~\ref{sec:method:training}.

\subsection{Geometry Toonification with StyleField}
\label{sec:method:geo_toonification}
To train a 3D generator $\Tilde{\generator}_{\rendererflag}$ capable of synthesizing artistic domain geometry, previous methods~\cite{Jin2022Dr3DA3,kim2022datid3d,lan2022e3dge,Abdal20233DAvatarGANBD,song2022diffusion} fine-tune the pre-trained $\rendererG$ with a target-domain dataset, which can be computationally expensive and could potentially deteriorate the original GAN latent space. To address this issue, we propose to establish a correspondence between the stylized NeRF $\nerf_{S}$ and the real-space NeRF $\nerf_{R}$. More specifically, we use a stylization field (the StyleField), $H_D$, to bridge the correspondence, such that $\Tilde{\generator}_{\rendererflag} (\point_S) = (\rendererG \circ H_D)(\point_S)$. 
As shown in Fig.~\ref{fig:framework}, given a stylized NeRF $\nerf_{S}:\mathbb{R}^3 \mapsto \mathbb{R}^4$ of the target style domain,
our goal is to estimate a 3D deformation residual, $H_D: \mathbb{R}^3 \mapsto \mathbb{R}^3 $, which maps $\nerf_{S}$ back to $\nerf_{R}$ via:
\begin{align}\label{eq:deformation_formulation}
    \nerf_{S} \xrightarrow{} \nerf_{R}: \point_R = (\point_S + H_D(\point_S)), \forall \point_{S} \in \nerf_{S},
\end{align}
where $H_D$ represents the residual 3D deformation $H_D(\point_S) = \Delta{\point_S}$ in the 3D space of the Stylized NeRF $\nerf_{S}$ and maps each 3D point $\point_{S} \in \nerf_{S}$ in the stylized space to its corresponding position in the real space $\nerf_{R}$.

To improve expressiveness, we extend $H_D$ as a conditional neural deformation field that outputs the offsets under the conditions of style and identity:
\begin{align}\label{eq:deformation_field}
    \nerf_{S} \xrightarrow{} \nerf_{R}: \point_R = \point_S + H_{D}(\point_S, \wcode_{S}, \wcode_R)
\end{align}
{where $\wcode_{S}$ is the style code that specifies the artistic domain, $\wcode_R$ is the instance code corresponding to $\nerf_{R}$ that represents the {identity} of the 3D face in the source domain. Both $\wcode_{S}$ and $\wcode_R$} serve as the holistic geometry indicators to guide the deformation.

We build $H_{D}$ as an MLP consisting of four \textsc{siren}~\cite{sitzmann2020siren} layers due to its superior high-frequency modeling capacity.
After all the points $\point_{S} \in \nerf_{S}$ are deformed to the real space, we generate the new feature map $\hat{\mathbf{F}}=\Tilde\generator_{\rendererflag}(\wcode_R,\wcode_S,\campose)$ for the input of $\decoderG$ and synthesize the high-resolution image with the geometry of the artistic domain.
{By introducing the 3D deformation module $H_D$, we no longer need to fine-tune pre-trained $\rendererG$ to achieve geometry deformation of the target domain. This greatly alleviates the parameters to optimize by 50\% and fully preserves the original GAN latent space. {Moreover, with $\wcode_{S}$ serving as the style condition, a single $H_D$ can support multiple styles, which further saves storage by 98.5\% compared to fine-tuning the whole model per style with $10$ styles.}}

\subsection{Texture Transfer with Adaptive Style Mixing}
\label{sec:method:tex_toonification}
Thanks to the StyleField $H_D$ that handles the geometry toonification within the cascade 3D GAN model, 
we only need to inject the artistic domain texture information into $\decoderG$ for texture stylization. 
Here, $\decoderG$ is a 2D style-based architecture, where image styles are effectively adjusted by AdaIN~\cite{huang_arbitrary_2017}.
Inspired by style mixing~\cite{karras2019style,karras_analyzing_2020,yang2022Pastiche} that controls the AdaIN parameters,
{we inject the texture information of the target style $\wcode_{S}$ by mixing the style parameters of $\decoderG$, as shown in Fig.~\ref{fig:framework}.}
To bridge the domain gap between the real space and target domain, we further add a lightweight MLP, $T$, for each layer of $\decoderG$ to adjust the style code $\wcode_{S}$. 
The adapted $T(\wcode_{S})$ and the instance code $\wcode_{R}$  are fused with a weight $w$ by weighted average, and sent to the affine transformation block of $\decoderG$ to obtain the final style parameters for AdaIN.
This mechanism allows us to model and control multi-domain textures with $T$ and $w$, without fine-tuning the original decoder $\decoderG$.

{Let $\Tilde{\generator}_{\decoderflag}$ denote $\decoderG$ with the adaptive style mixing. The image generation process with domain adaptation given $w$ and $\wcode_S$ can be formulated as $\imagePredicted=\Tilde{\generator}_{\decoderflag}(\hat{\mathbf{F}}, \wcode_R,\wcode_S, w)$.}

\subsection{Training}
\label{sec:method:training}

\heading{Data Preparation}
We follow Sim2Real~\cite{Wood2021FakeIT,zhang21datasetgan,lan2022e3dge} to generate paired data for training. 
Specifically, to generate the training corpus for each iteration of the training process, we pre-calculate a set of real space NeRFs $\nerf_R$ with corresponding latent codes $\wcode_R$ and rendered images $\image_R$.
To generate pair-wise stylized ground truths, we stylize the rendered image with existing 2D toonification models to obtain the target ground truth $\image_S$.
Here, to validate our method's performance on multi-style domain adaptation, we adopt the exemplar-based DualStyleGAN~\cite{yang2022Pastiche} as our 2D toonification model, since it supports hundreds of diverse styles, such as Cartoon, Pixar, and Caricature.

Finally, we define $\datasetsample = \{\image_R, \wcode_R,\wcode_S, \image_S\}$ as a training set for \METHODNAME~{with \{$\wcode_R$, $\image_R$, $\wcode_S$\} to serve as the training inputs and \{$\image_S$\} is the set of training ground truth.} 
$\image_R \in \datasetsample$ is drawn i.i.d from distribution $P(G(\zcode, \campose))$ where $\zcode \sim \mathcal{N}(0,1)$ and $\campose$ is the camera pose distribution of pre-trained 3D GAN $G$.
Some samples are shown in Fig.~\ref{fig:training_sample}.

\heading{Reconstruction Loss}
Here we use the $\mathrm{LPIPS}$ loss~\cite{zhang2018perceptual} to evaluate stylized image quality:
\begin{equation}
\vspace{-1.25mm}
    \loss_{\text{Rec}}\left ( \datasetsample \right ) =
    {\mathbb{E}}_{\datasetsample}
    \Bigg[
    || P(\imagePredicted) - P(\image_S) ||_2
    \Bigg]
    ,
\end{equation}
where $P(\cdot)$ denotes the perceptual feature extractor.

\input{tabs_and_figs/fig_training_samples}

\heading{Smoothness Regularization}
To encourage the smoothness of stylization field deformation offsets and reduce spatial distortion,
a smoothness regularization is included to regularize $H_D$.
Here we penalize the norm of the Jacobian matrix  $\mathbb{J}_{H_D} = \nabla {H_D}$ of the deformation field~\cite{park2021nerfies}
to ensure the learned deformations are physically smooth:
\vspace{-1.25mm}
\begin{equation}
\begin{aligned}
\label{eq:smoothness}
    &\mathcal{L}_{\mathrm{Elastic}} = 
    \mathrm{ReLU}(
    \norm{
   \nabla H_D(\point_S, \wcode_S, \wcode_R) 
    }_{2}^{2}-\epsilon),
\end{aligned}
\end{equation}

\noindent
where $\epsilon$ is the slack parameter for the smoothness regularization. We set $\epsilon=0.1$ for all the experiments.

\heading{Adversarial Training}
Additionally, we apply a non-saturating adversarial loss $\loss_{\mathrm{Adv}}$~\cite{karras2019style} to bridge the domain gap of toonified results.

In summary, the overall loss function is defined as
\vspace{-1.25mm}
\begin{equation*}
% \small
    \loss = \loss_{\mathrm{Rec}} + \lossweight_{\mathrm{Elastic}}\loss_{\mathrm{Elastic}} + \lossweight_{\mathrm{Adv}}\loss_{\mathrm{Adv}}, %  + \loss_{\mathrm{D}} + \loss_{\mathrm{R1}}.
\end{equation*}
% \vspace{-1.25mm}
where we set $\lossweight_{\mathrm{Adv}}=0.05$ and $\lossweight_{\mathrm{Elastic}}=0.01$ in all the experiments.

%% file: tabs_and_figs/fig_framework.tex
% !TEX root = ../main.tex

\begin{figure*}[h!]
\begin{center}
\vspace{-4mm}
  \includegraphics[width=1\linewidth]
  {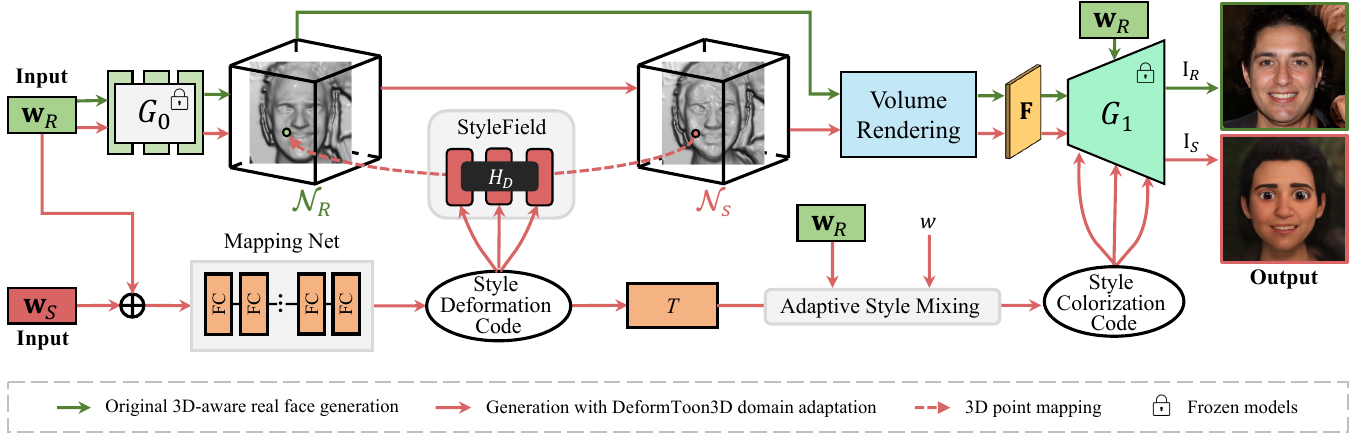}
\end{center}
\vspace{-4mm}
  \caption{
  % \cavan{give description. Many symbols described in the text are missing. Don't use green font, it is hard to see. Use the same font for symbols as in the main text. If there is empty horizontal space, show some examples of input and output. }
  \textbf{\METHODNAME{} framework.} Given sampled instance code $\wcode_R$ and style code $\wcode_S$ as conditions, \METHODNAME{} first deforms a point from the style space $\nerf_S$ to the real space $\nerf_R$, which achieves geometry toonification without modifying the pre-trained $\rendererG$. Afterwards, we leverage adaptive style mixing with weight $w$ to inject the texture information of the target domain into the pre-trained $\decoderG$ for texture toonification.
  % The original 3D-aware real face generation is shown in "GREEN" line, and the \METHODNAME{} domain adaptation generation is shown in "RED" line. 
  Both pre-trained generators $\rendererG$ and $\decoderG$ are kept frozen.
  }
  \vspace{-3mm}
\label{fig:framework}
\end{figure*}

%% file: tabs_and_figs/fig_training_samples.tex
% !TEX root = ../main.tex
\begin{figure}[h!]
\begin{center}
  % \vspace{-1mm}
  \includegraphics[width=1\linewidth]{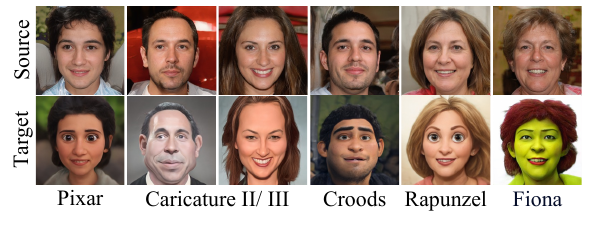}
\end{center}
\vspace{-5mm}
  \caption{\textbf{Training Samples}. We show the source image (row $1$) and the stylized images of the target domain (row $2$) for training supervisions.}
\label{fig:training_sample}
\vspace{-4mm}
\end{figure}

%% file: sections/4_experiment.tex
% !TEX root = ../main.tex

\input{tabs_and_figs/fig_domain_adaptation}

\section{Experiments}
\label{sec:experiment}

\heading{Datasets} 
We mainly focus on the human face domain and use synthesized data for the whole training.
We follow the data synthesis procedure mentioned in Sec.~\ref{sec:method:training} for generating training corpus and adopt DualStyleGAN~\cite{yang2022Pastiche} as the 2D stylization model to infer paired toonified image.
We generate images with the following 10 styles: Pixar, Comic, Slam Dunk, The Croods, {Fiona (Shrek)}, Rapunzel~\text{(Disney Princess)}, Hiccup Horrendous Haddock III~\text{(How To Train Your Dragon)}, and three different caricature styles.
{We use StyleSDF~\cite{orel2021stylesdf} pre-trained on FFHQ~\cite{karras2019style} as our 3D generator.} 
For evaluating toonification on real-world images, we evaluate on CelebA-HQ~\cite{karras2018progressive}.

\heading{Implementation Details} 
In all the experiments, we set the learning rate to $5\times10^{-4}$.
We adopt Adam~\cite{Kingma2015AdamAM} optimizer to train the toonification models.
We train the \METHODNAME~for 100 epochs. To expedite the training process, we disable the adversarial loss for the initial 50 epochs. The training takes approximately 24 hours using 8 Tesla V100 GPUs with a batch size set to 16.
For adaptive style mixing, {$w=1$ is fixed during training and could be manually selected from $[0,1]$ to achieve texture interpolation during inference.}
More implementation details and experiment results are included in the supplementary material.

\heading{Baselines}
Here we design three baselines for extensive evaluations. 
The first is CIPS-3D~\cite{zhou2021CIPS3d}, which naively fine-tunes the super-resolution $\decoderG$ for view-consistent toonification.
The second is E3DGE~\cite{lan2022e3dge}, which fine-tunes both $\rendererG$ and $\decoderG$ independently for true-3D toonification.
Another prominent method is to leverage directional CLIP loss~\cite{Gal2021StyleGANNADACD,alanov2022hyperdomainnet,kim2022datid3d} for adapting a pre-trained style-based generator. Here we extend StyleGAN-NADA~\cite{Gal2021StyleGANNADACD} to 3D GAN and employ image CLIP directional loss for the evaluation.
The baseline models are trained on the same dataset as \METHODNAME{}, with each fine-tuning process applied to a single style. In contrast, \METHODNAME{} is capable of accommodating all styles within a single model.

\subsection{Comparisons with Baselines}
\heading{Qualitative Results}
We show the qualitative comparisons against the baselines in Fig.~\ref{fig:qualitative_comparisons}. \METHODNAME{} fully captures the characteristics of the target domain with consistent identity preservations. CIPS-3D only provides texture-wise toonification and ignores geometry deformation. E3DGE suffers from mode collapse and tends to lose identity. StyleGAN-NADA fails to capture the characteristics of the target domain. Our method produces high-quality toonification with consistent identity preservations.

\input{tabs_and_figs/tab_id_fid}

\heading{Quantitative Results}
To evaluate the fidelity and quality of toonification, we compare their identity preservation($\mathrm{IP}$)~\cite{deng2018arcface} and $\mathrm{FID}$ respectively in Tab.~\ref{ch:deformtoon3d:tab:id_fid}.
In terms of $\mathrm{IP}$, \METHODNAME{} outperforms all baseline methods across the 10 styles provided, which underscores the benefits of retaining the 3D generator.
In terms of $\mathrm{FID}$, CIPS-3D~\cite{zhou2021CIPS3d} only fine-tunes $\decoderG$ and achieves worse performance compared to E3DGE~\cite{lan2022e3dge}, which fine-tunes the whole generator. 
StyleGAN-NADA achieves the worst $\mathrm{FID}$ performance, which we attribute to the challenges of directly adopting 2D CLIP-based supervision on 3D GANs.
A detailed breakdown by individual styles is available in the supplementary material.

\input{tabs_and_figs/tab_user_study}

\heading{User Study}
For the user preference study, we collected 2400 votes to select the preferred rendering results in terms of shape stylization, appearance stylization, identity preservation, and overall performance. As shown in Tab.~\ref{ch:deformtoon3d:tab:user_study_toonification}, the proposed method gives the most preferable results despite the fact that the baseline methods are trained on a single style.

\input{tabs_and_figs/tab_storage_cost}

\heading{Storage Cost Comparison} 
We detail the storage costs in Table~\ref{tab:storage_cost}. Thanks to our unique design that disentangles geometry and texture, our method requires no fine-tuning and fewer parameters for training. In a single-style scenario, our method reduces storage needs by 85\%. 
This advantage becomes even more pronounced with the multi-style version of $H_D$, achieving a storage saving of 98.5\% in a 10-style scenario. This makes our approach particularly feasible for potential mobile applications.

\subsection{Applications}
To demonstrate the generality of full GAN latent space preservation, we show that \METHODNAME~could be easily extended to a series of downstream applications proposed for the original pre-trained GAN, including inversion, editing, and animation.
To further validate \METHODNAME's unique geometry-texture decomposed toonification, we show results of flexible toonification style control.

% \subsection{Latent Space Exploration}
\input{tabs_and_figs/fig_editing}
\input{tabs_and_figs/fig_real_image_stylization}
\heading{Inversion and Editing}
With fully preserved 3D generative prior, \METHODNAME{} could directly adopt pre-trained 3D GAN inversion framework and latent editing directions for semantic-aware editing over the toonified portrait.
Here we adopt E3DGE~\cite{lan2022e3dge} for 3D GAN inversion and show the inversed results of the real image in Fig.~\ref{fig:inversion}. With fully preserved GAN latent space, \METHODNAME{} could be applied to real images over multiple styles with high quality.
We also include the editing results of diverse styles in Fig.~\ref{fig:editing}.
As can be seen, our method produces consistent editing results and fully preserves the identity and the style of the target domain, regarding both the geometry and texture. 

\input{tabs_and_figs/fig_animation}
\heading{Animatable Toonification}
Inspired by the success of 2D method~\cite{tewari2020stylerig} that obtains a rig-like control over StyleGAN-generated 2D faces, 
we propose a straightforward pipeline that aligns 3DMM~\cite{blanz1999morphable} parameters with pre-trained 3D GAN latent space.
Specifically, we train two MLP $\mathcal{F}_{\ww}:\mathbb{R}^{\abs{\wspace}} \rightarrow \mathbb{R}^{\abs{\mmspace}}$ and $\mathcal{F}_{\mm}:\mathbb{R}^{\abs{\mmspace}} \rightarrow \mathbb{R}^{\abs{\wspace}}$ to learn the bi-directional mapping between 3DMM parameter space $\mmspace$ and 3D GAN $\wspace$ space. 
To impose 3DMM-based control, given a latent code $\ww$, we first infer its 3DMM parameters $\Tilde{\mm} = \mathcal{F}_{\ww}(\ww)$ and reconstructed 3DMM code $\Tilde{\ww} = \mathcal{F}_{\mm}(\mm_{\ww})$. After imposing 3DMM-based editing $\Tilde{\mm}_{\Delta} = \Tilde{\mm} + \mm_{\Delta}$, we infer the corresponding edited $\wspace$ code $\Tilde{\ww}_{\Delta} = \mathcal{F}_{\mm}(\Tilde{\mm}_{\Delta})$. The final result is synthesized from $\hat{\ww} = \ww + (\Tilde{\ww}_{\Delta} - \Tilde{\ww})$.
The whole framework is trained in a self-supervised manner with cycle consistency regularizations.
Please refer to the supplementary material for more technical details and animation results.

We show the animated toonification results in Fig.~\ref{fig:animation}. 
Here, we extract the 3DMM parameters from a driving video~\cite{guo2021adnerf} using a pre-trained predictor~\cite{deng2019accurate}. The expression dimension of extracted parameters are injected to the sampled $\wcode_R$ using the procedure described above. Four styles of animated toonified results are included. As can be seen, our designed animation pipeline could accurately drive the identity both in the real space (row $2$) and the style spaces (row $3-6$). The reenacted expressions are natural and abide with the driving input, which validates the generality of our method.

\input{tabs_and_figs/fig_style_swap}

\input{tabs_and_figs/fig_style_interp}

\heading{Toonification Style Control}
Due to the unique geometry-texture decomposition design, \METHODNAME{} offers flexible style control.
\textbf{First}, we achieve style degree control and show the results in Fig.~\ref{fig:style_degree_control}.
Geometry-wise, since $H_D$ outputs the 3D deformation offsets $\Delta{\point_S}$, we simply interpolate the offsets with $\tau*\Delta{\point_S}$ where $\tau=0$ represents an identical mapping of the real space.
Texture-wise, we rescale the style mixing weight $w$ of $\decoderG$, where $w=0$ preserves the color of the real space images.
\textbf{Second}, due to the geometry-texture disentangled property, \METHODNAME{} naturally supports the toonification of shape and texture only. As shown in Fig.~\ref{fig:style_swap}, we achieve the geometry-texture swap between multiple styles, where the geometry of one style could be combined with the texture of another style. This could not be achieved by all previous methods and opens up broader potential downstream applications.

\input{tabs_and_figs/tab_ablation}
\subsection{Ablation Study}
We ablate the effectiveness of our design choices in Tab.~\ref{tab:ablation}.
\textbf{1) Instance code condition:} By removing instance code $\wcode_R$, StyleField tends to learn similar offsets for different identities, whereas adopting $\wcode_R$ as the instance deformation conditions facilitates better toonification.
\textbf{2) StyleField architecture:} We replace the \textsc{siren} architecture with an MLP network as defined by D-NeRF~\cite{pumarola_d-nerf_2020} and observe the significant performance drop. This demonstrates the representation power of \textsc{siren} deformaiton field. 
\textbf{3) Adversarial training:} We also validate the effectiveness of adversarial loss, which brings noticeable improvement regarding both $\mathrm{FID}$ and identity similarity, respectively.

\subsection{Limitations} 
As shown in Fig.~\ref{ch:deformtoon3d:fig:failure_case} pertaining to the style ``Comic" and ``Slam Dunk",
though DeformToon3D produces reasonable texture-wise stylization, the corresponding geometry still has noticeable artifacts. 
The proposed StyleField implicitly learns the correspondence between the paired data from the real space and the style space. Such correspondence is easier to learn with information cues such as illumination from the 3D-ish styles like Pixar or noticeable keypoints from caricature styles, but harder for styles with limited information cues like Comic. 

\input{tabs_and_figs/fig_failure_case}

%% file: tabs_and_figs/fig_domain_adaptation.tex
% !TEX root = ../main.tex

\begin{figure*}[h!]
\vspace{-3mm}
\begin{center}
  \includegraphics[width=1\linewidth]{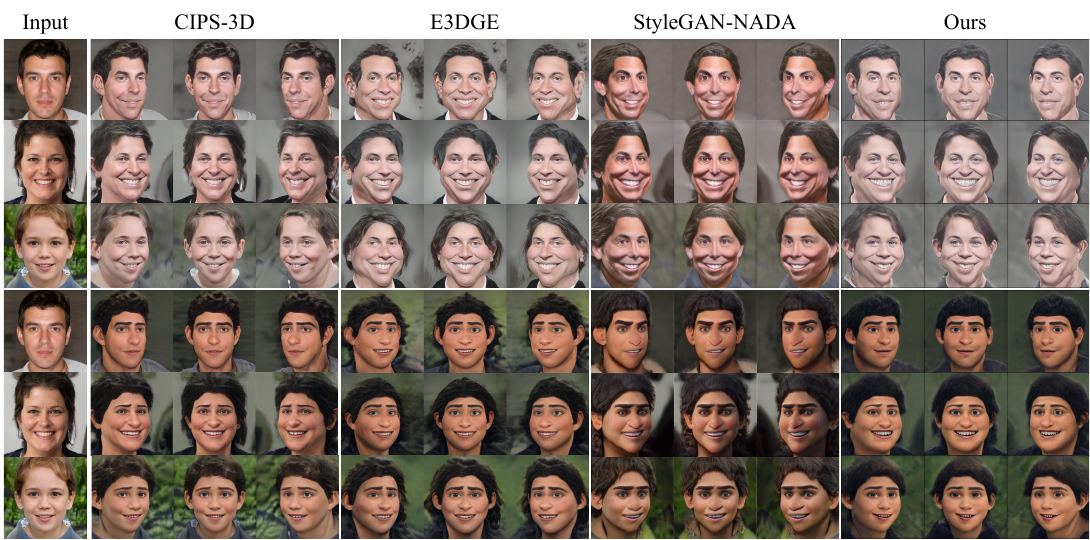}
\end{center}
\vspace{-4mm}
  \caption{\textbf{Qualitative comparisons with baseline methods.} \METHODNAME{} produces better performance against all baselines regarding toonification fidelity, diversity and identity preservation.
  }
\vspace{-3mm}
\label{fig:qualitative_comparisons}
\end{figure*}

%% file: tabs_and_figs/tab_id_fid.tex
% !TEX root = ../main.tex
\begin{table}[h]
\small
\vspace{-1mm}
\begin{center}
\caption{\textbf{Quantitative evaluation over 10 styles.} \METHODNAME{} achieves the best identity preservation (IP) and $\mathrm{FID}$.}
\vspace{-2mm}
\label{ch:deformtoon3d:tab:id_fid}
\begin{tabular}{l|c|c|c|c}
\hline
& CIPS-3D  &  E3DGE & NADA & Ours   \\ 
\hline  
$\mathrm{IP}$$\uparrow$		& 0.681	& 	0.707	& 0.535	& 		\textbf{0.781}  \\ 
\hline  
$\mathrm{FID}$$\downarrow$ &	50.6 &	34.0	 & 59.3 &	\textbf{27.6}   \\ 
\hline
\end{tabular}
\end{center}
\vspace{-4mm}
\end{table}

%% file: tabs_and_figs/tab_user_study.tex
% !TEX root = ../main.tex

\setlength{\tabcolsep}{4.5pt}
\begin{table}[h]
\small
\vspace{-1mm}
\begin{center}
\caption{\textbf{User preference study.}}
\label{ch:deformtoon3d:tab:user_study_toonification}
\vspace{-2mm}
\begin{tabular}{l|c|c|c|c}
\hline
   & CIPS-3D  &  E3DGE & NADA & Ours   \\ 
\hline
Shape  &	 20.8\%	& 	17.4\%	& 3.4\% & \textbf{58.4\%}	\\ 
\hline
Appearance & 16.1\% & 16.4\% & 3.4\%  &\textbf{ 64.1\%}	  \\
\hline
Identity  & 23.1\%	 & 7.1		& 5.4  & \textbf{64.4\%}	\\
\hline
Overall  & 21.8\%	 & 9.9\%		 & 2.7\%  & \textbf{65.6\%}	\\
\hline
\end{tabular}
\end{center}
\vspace{-4mm}
\end{table}

%% file: tabs_and_figs/tab_storage_cost.tex
% !TEX root = ../main.tex

\setlength{\tabcolsep}{4.5pt}
\begin{table}
\small
\vspace{-1mm}
\caption{\textbf{Storage cost comparison}. Values are averaged across $10$ styles and shown in $\mathrm{MB}$. 
\METHODNAME{} achieves a considerably more efficient storage footprint against the baselines.
} 
\label{tab:storage_cost}
\vspace{-5mm}
\begin{center}
\begin{tabular}{l|c|c}
\hline
Methods  & Trainable Params$\downarrow$ &  Model Storage$\downarrow$ \\ 
\hline
CIPS-3D~\cite{zhou2021CIPS3d} & 5.81  & 59.93 	\\
E3DGE~\cite{lan2022e3dge}  & 7.64  & 76.4  \\
StyleGAN-NADA~\cite{Gal2021StyleGANNADACD} & 7.64  & 76.4	\\
% Ours-S & 3.82 & 45.84    \\ % 3.82x10+ 
\hline
Ours (single-style) & \textbf{3.82} & \textbf{11.46}  \\
\hline
\end{tabular}
\end{center}
\vspace{-6mm}
\end{table}
% \vspace{-4mm}

%% file: tabs_and_figs/fig_editing.tex
% !TEX root = ../main.tex

\begin{figure}[h!]
\vspace{-2mm}
\begin{center}
  \includegraphics[width=1\linewidth]{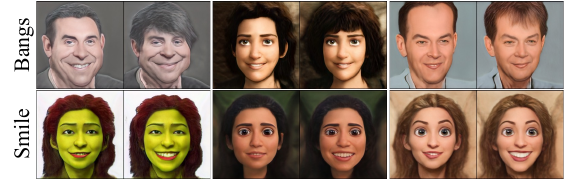}
\end{center}
\vspace{-4mm}
  \caption{\textbf{Editing of toonified results.} We show two attribute editing results over six styles. In row $1$, we add bangs to the male identities, and in row $2$, we add ``Smile'' to female identities. The edited results fully preserve the identity and abide by the style of the target domain.}
\label{fig:editing}
\vspace{-3mm}
\end{figure}
% \newpage

%% file: tabs_and_figs/fig_real_image_stylization.tex
% !TEX root = ../main.tex
\begin{figure*}[h!]
\vspace{-3mm}
\begin{center}
  \includegraphics[width=1\linewidth]{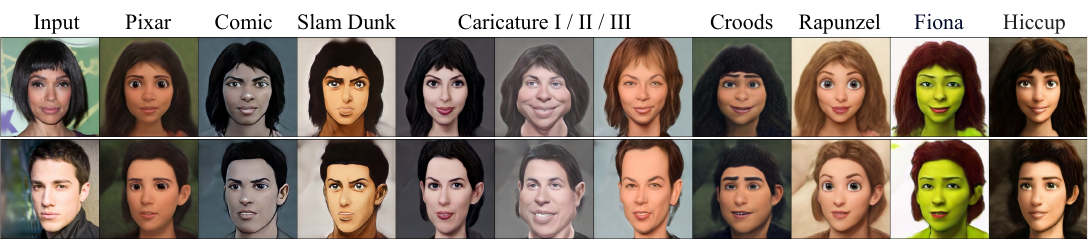}
\end{center}
  \vspace{-4mm}
  \caption{\textbf{\METHODNAME{} on the real images.} Our method enables multiple styles toonification with a single model, where both the texture and the geometry matches the target domain.}
  \vspace{-4mm}
\label{fig:inversion}
\end{figure*}

%% file: tabs_and_figs/fig_animation.tex
% !TEX root = ../main.tex

\begin{figure}[h!]
\begin{center}
  \includegraphics[width=0.95\linewidth]{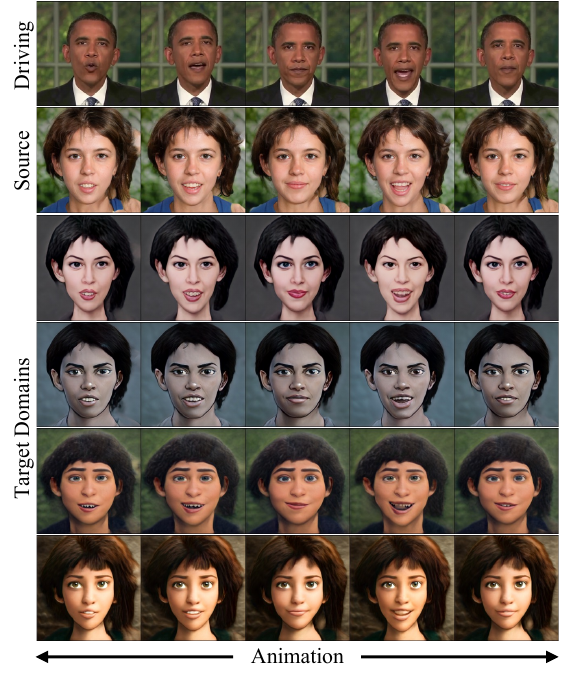}
\end{center}
  \vspace{-6mm}
  \caption{
  \textbf{Animation of stylized results.} We drive the real space images (row 2) with an input video (row 1). Since \METHODNAME{} fully preserves the pre-trained GAN latent space, the driving direction of the real space could be directly applied to the style space (row $3-6$). The expression on of animated toonified identities fully abide with the driving frames without affecting the toonification performance of target domain.}
  \vspace{-4mm}
\label{fig:animation}
\end{figure}

%% file: tabs_and_figs/fig_style_swap.tex
% !TEX root = ../main.tex
\begin{figure}[h!]
\vspace{-3mm}
\begin{center}
  \includegraphics[width=0.9\linewidth]{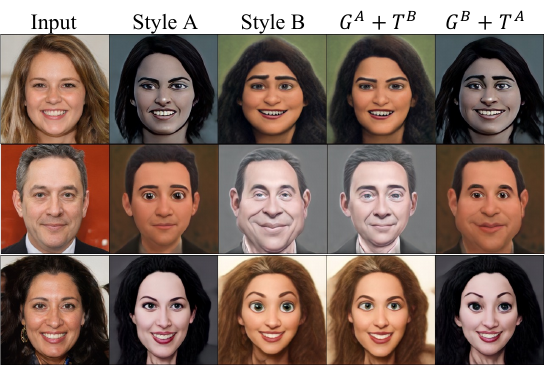}
\end{center}
\vspace{-3mm}
  \caption{\textbf{Style swap.} Besides style interpolation, \METHODNAME{} supports style swap of geometry and texture only across two styles. As shown here, given the real space input (col $1$) and the toonification results of two spaces (col $2-3$), \METHODNAME{} could swap the geometry $G$ and texture $T$ of two styles independently (col $4-5$), which cannot be achieved by previous methods.}
\vspace{-1mm}
\label{fig:style_swap}
\end{figure}

%% file: tabs_and_figs/fig_style_interp.tex
% !TEX root = ../main.tex

\begin{figure}[h!]
\vspace{-2mm}
\begin{center}
  \includegraphics[width=0.95\linewidth]{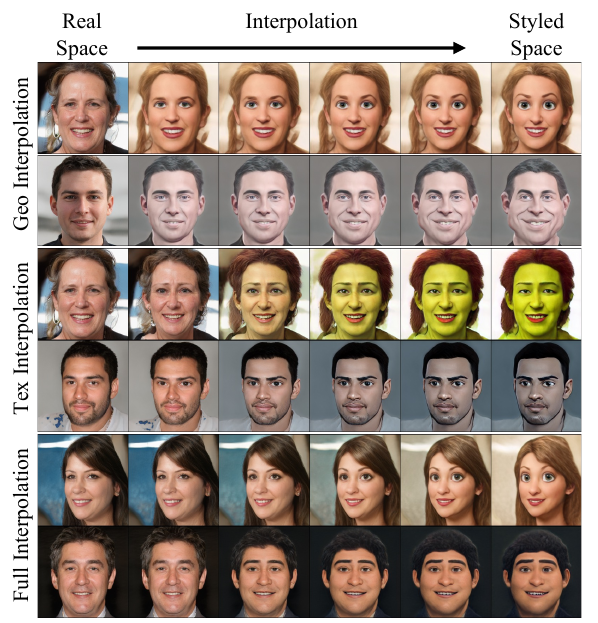}
\end{center}
\vspace{-4mm}
  \caption{\textbf{Style degree control.} Thanks to the geometry-texture toonification decomposition, \METHODNAME{} can specify geometry, texture, and full style control.}
\vspace{-2mm}
\label{fig:style_degree_control}
\end{figure}

%% file: tabs_and_figs/tab_ablation.tex
% !TEX root = ../main.tex
\setlength{\tabcolsep}{8pt}
\begin{table}
\small
\vspace{-1mm}
\caption{Ablation study.}
\label{tab:ablation}
\vspace{-4mm}
\begin{center}
\begin{tabular}{l|c|c}
\hline
Method    &  $\mathrm{IP}$$\uparrow$   & $\mathrm{FID}$$\downarrow$  \\ 
\hline
w/o $\wcode_R$ condition & 0.735 &  33.4  \\
MLP as $H_D$ & 0.746 & 31.8  \\
w/o $\loss_{\mathrm{Adv}}$ & 0.769 & 29.9  \\
\hline
Ours & \textbf{0.781} & \textbf{27.6}    \\ 
\hline
\end{tabular}
\end{center}
\vspace{-8mm}
\end{table}

%% file: tabs_and_figs/fig_failure_case.tex
% !TEX root = ../main.tex

\begin{figure}[h!]
\begin{center}
\vspace{-2mm}
  \includegraphics[width=1\linewidth]{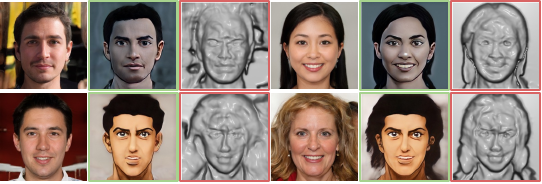}
\end{center}
\vspace{-4mm}
  \caption{\textbf{Failure cases.}}
\label{ch:deformtoon3d:fig:failure_case}
\vspace{-4mm}
\end{figure}

%% file: sections/5_conclusion.tex
% !TEX root = ../main.tex
\section{Conclusion and Future Work}
In this paper, we propose a novel 3D toonification framework \METHODNAME{} for a fine-tuning free, geometry-texture decomposed 3D face toonification. 
We fully exploit the hierarchical characteristics of 3D GAN and introduce a StyleField to handle 3D geometry toonification of $\rendererG$, along with adaptive style mixing that injects texture information into $\decoderG$.
Our method achieves high-quality toonification on both geometry and texture, outperforming existing methods. Thanks to the preservation of the 3D generative prior, \METHODNAME{} facilitates a range of downstream applications.

As a pioneering effort in this field, we believe this work will inspire future works on free 3D toonification.
First, to mitigate the geometry-texture ambiguity present in certain styles, introducing re-lighting during training could serve as a potential solution~\cite{pan2021shadegan}.
Second, a more flexible training paradigm could directly guide the 3D toonification process with a pre-trained vision-language model~\cite{Rombach2021HighResolutionIS}. 
Third, future research could focus on integrating a comprehensive 3D animation pipeline~\cite{Bergman2022GenerativeNA,sun2023next3d} into the toonification process.
Moreover, the potential applicability of DeformToon3D to other 3D GANs~\cite{Chan2021EG3D} and shapes beyond human faces, such as full-body settings, is worth investigating.

%% file: sections/6_appendix.tex
% !TEX root = ../main.tex

\section{Background}
Since recent 3D-aware image generative models are all based on neural implicit representations, especially NeRF~\cite{mildenhall2020nerf}, here we briefly introduce the NeRF-based 3D representation and more StyleSDF details for clarification.

\heading{NeRF-based 3D Representation}
NeRF~\cite{mildenhall2020nerf} proposed an implicit 3D representation for novel view synthesis. 
Specifically, NeRF defines a scene as $\{\bm{c},\sigma\} = F_{\Phi}(\point, \view)$, 
where $\point$ is the query point, $\view$ is the viewing direction from camera origin to $\point$, $\bm{c}$ is the emitted radiance (RGB value), 
$\sigma$ is the volume density. 
To query the RGB value $C(\bm{r})$ of a point on a ray $\bm{r}(t)=\bm{o}+t\view$ shoot from the 3D coordinate origin $\bm{o}$, 
we have the volume rendering formulation,
\begin{equation}
\label{eq:volum_render}
C(\bm{r}) = \int_{t_n}^{t_f}T(t)\sigma(\bm{r}(t))\bm{c}(\bm{r}(t), \bm{v})dt,
\end{equation}
where $T(t)=\text{exp}(-\int_{t_n}^{t}\sigma(\bm{r}(s))ds)$
is the accumulated transmittance along the ray $\bm{r}$ from $t_n$ to $t$. $t_n$ and $t_f$ denote the near and far bounds. 

\heading{Hybrid 3D Generation}
In hybrid 3D generation~\cite{orel2021stylesdf,Chan2021EG3D,gu2021stylenerf}, the intermediate feature map is calculated by replacing the color $\bm{c}$ with feature $\mathbf{f}$, namely $\mathbf{F}(\mathbf{r}) = \int_{t_n}^{t_f}T(t)\sigma(\mathbf{r}(t))\mathbf{f}(\mathbf{r}(t),\bm{v})dt$.
Then, a StyleGAN~\cite{karras2019style,karras2020analyzing}-based decoder upsamples $\mathbf{F}$ into high-resolution images with high-frequency details.

\heading{SDF and Radiance-based Geometry Representation}
The intermediate geometry representation of $\rendererG$ diversifies the characteristics of different 3D GANs.
Specifically, StyleSDF~\cite{orel2021stylesdf} uses $\rendererG$ to predict the signed distance $\dist(\point)=\rendererG(\wcode,\point)$ between the query point $\point$ and the shape surface, where the density function $\sigma(\point)$ can be transformed from $\dist(\point)$~\cite{orel2021stylesdf,wang2021neus,yariv2021volume} for volume rendering~\cite{mildenhall2020nerf}.
The incorporation of SDF leads to higher-quality geometry in terms of expressiveness view consistency and clear definition of the surface.

In this paper, we mainly adopt StyleSDF~\cite{orel2021stylesdf} due to its high-quality geometry surface and high-fidelity texture.
In StyleSDF, the Sigmoid activation function $\sigma$ is replaced by $\sigma(\point) = K_{\alpha} \left( \dist(\point) \right) =  \text{Sigmoid}\left(-\dist(\point)/\alpha\right)/\alpha$,
where $\alpha$ is a learned parameter that controls the tightness of the density around the surface boundary.

\section{Implementation Details}
\heading{CIPS-3D Baseline} 
Following CLIPS-3D~\cite{zhou2021CIPS3d}, we fine-tune $\decoderG$ of StyleSDF on the toonified images with identical optimization parameters in the official implementation. The fine-tuning time for one style costs $10$ V100 minutes.

\heading{E3DGE Baseline} 
Following E3DGE~\cite{lan2022e3dge}, we first fine-tune $\rendererG$ for $400$ iterations with batch size $24$, and further fine-tune $\decoderG$ for $400$ iterations with batch size $8$. All hyper-parameters are left unchanged with the official StyleSDF~\cite{orel2021stylesdf} implementation.
The overall fine-tuning time for one style costs around 30 minutes on a single V100 GPU.

\heading{StyleGAN-NADA Baseline} 
We reproduce StyleGAN-NADA~\cite{Gal2021StyleGANNADACD} on StyleSDF with the following modifications. For $\rendererG$ optimization, we fix the pre-trained mapping network, affine code transformations, view-direction MLP, color-prediction MLP, and density-prediction MLP.  
For $\decoderG$ optimization, we follow the original implementation and fine-tune all weights except toRGB layers, affine code transformations, and mapping network.
The $k$ layers to optimize are also selected adaptively using StyleCLIP global loss.
Other hyper-parameters and training procedures are left unchanged.
The whole optimization costs around $5$ minutes on a single V100 GPU.

\subsection{Additional Method Details}
\heading{StyleField}
Given the instance code $\wcode$, style code $\code_S$, we concatenate them along the channel dimension and send them into a $4$-layer mapping network~\cite{karras2019style}.
The mapping network first maps $\wcode \oplus \code_S$ to a set of modulation signals $\{\bm{\beta}, \bm{\gamma}\}$, where $\bm{\beta}=\{\beta_i\}, \bm{\gamma}=\{\gamma_i\}$.
To associate the given codes to the corresponding deformation,
the modulation signals will be injected into the MLP network, serving as FiLM conditions \cite{perez2018film,dumoulin2018feature-wise,sitzmann2020siren} to modulate its features at different layers as $\vf_{i+1}=\sin (\gamma_{i}\cdot(\mW_{i}\vf_i+\vb_i)+\beta_i)$.
To support multi-style code, we associate each style index with a learnable embedding. During inference, we pass in the corresponding style index to retrieve the style embedding for conditional deformation.

\input{tabs_and_figs/fig_3dmm_animation}
\heading{Animatable Stylized Portrait Training Details} 
We show the overall 3DMM animation inference pipeline in Fig.~\ref{fig:3dmm_animation}.
Specifically, we train the whole framework in a self-supervised manner. In each iteration, we synthesize a batch of pose images $\image = G(\ww)$, where $G = \decoderG \circ \rendererG$. For 3DMM supervision, we leverage the state-of-the-art 3DMM predictor~\cite{deng2019accurate} to infer the pseudo ground-truth 3DMM parameter $\mm_{\text{GT}}$. With the synthesized training corpus, we reconstruct the input codes $\hat{\mm} = \mathcal{F}_{\ww}(\ww)$ and $\hat{\ww} = \mathcal{F}_{\mm}(\mm_{\text{GT}})$ and impose MSE reconstruction loss. We further render the reconstructed code $\Tilde{\image} = G(\hat{\ww})$ and $\Tilde{\image}_{\mmspace} = \mathrm{DFR}(\hat{\mm})$, where $\mathrm{DFR}$ is a differentiable render~\cite{ravi2020pytorch3d} that renders the reconstructed 3DMM mesh to image.
The rendered images are supervised with corresponding loss~\cite{deng2019accurate}, which yields better performance in our observations.

This training objective shall guarantee plausible 3DMM editing using the procedure stated in the main context. 
However, in practice, we find the training is unstable and predicted codes are not disentangled well during inference. We make the following modifications to the overall training pipeline and improve the editing performance:

{First}, we observe that the 3DMM head pose parameters, including a head rotation $\mathbf{R}\in SO(3)$ and translation $\mathbf{t}\in\mathbb{R}^3$ parameters, are in contradict with the pose control of 3D GAN. This deteriorates the disentanglement of learned codes and destabilizes the training since the predicted 3DMM codes $\Tilde{\mm}$ must contain accurate head pose to minimize the reconstruction loss with $\mathrm{DFR}(\mm)$. To address this issue, we mask out the head rotation $\mathbf{R}$ and translation $\mathbf{t}\in\mathbb{R}^3$ dimension in all 3DMM parameters with the binary mask.  This enforces all the 3DMM images $\image_{\mmspace}$ to be rendered from frontal pose and encourages the networks to focus on facial expression $\boldsymbol\delta\in\mathbb{R}^{64}$ alignment between $\wspace$ and $\mmspace$.

{Second}, to further impose identity preservation and bijective mapping between two spaces, we introduce cycle training~\cite{CycleGAN2017} which regularizes $\ww \approx \mathcal{F}_{\ww}( \Tilde{\mm} )$ and  $\mm \approx \mathcal{F}_{\mm}(\Tilde{\ww})$. The cycle loss is also imposed on the image space.

{Third}, to imitate the inference pipeline, in each training iteration, we randomly shuffle the expression dimension of all the 3DMM code $\mm_{\text{GT}}$ within a batch and generate a new set of codes $\Tilde{\mm}_{\text{GT}}$. The rendered image from $\mathcal{F}_{\mm}(\Tilde{\mm}_{\text{GT}})$ shall maintain the same identity with $\image$ with identical pose of $\mathrm{DFR}(\Tilde{\mm}_{\text{GT}})$. We impose the identity preservation loss~\cite{deng2018arcface} and landmark loss over the rendered 3DMM image~\cite{deng2019accurate} as supervisions. This strategy further reduces the domain gap between training and inference and further improves the final editing performance.

{Fourth}, we further leverage the style-based hierarchical structure within StyleSDF and reduces the attribute entanglement. Specifically, rather than using the edited $\wspace$ code $\hat{\ww}_{\Delta}$ for all the style layers in $G$, we conduct layer-wise editing effect analysis and find that only the first $2$ layers of $\rendererG$ will handle the expression-relevant information of the synthesized image. Using the edited code for later layers will result in other attributes editing, \eg, adding glasses or changing the hair structure. Therefore, we leave the remaining $7$ layers of $\rendererG$ and all $10$ layers in $\decoderG$ unchanged and only use the edited code for the top $2$ $\rendererG$ layers. This yields better disentanglement during the 3DMM-controlled style editing.

Training-wise, we adopt identical MLP architecture from PixelNeRF~\cite{yu2021pixelnerf} to implement both $\mathcal{F}_{*}$ networks and adopt a batch size of $4$ with learning rate $5\times10^{-4}$ during the optimization. The networks are trained for $50,000$ iterations, which costs around 2 days on a single V100 GPU.
Please refer to the released code for more details.

\subsection{Additional Ablation Study}
The robustness of the number of style codes is ablated in Table~\ref{ch:deformtoon3d:tab:ablate_styles}. Experiments are conducted with 1, 2, and 5 styles per model so that under each setting the 10 styles can be evenly divided into different runs for the sake of comparison.

\input{tabs_and_figs/tab_ablation_styles}

The effectiveness of the elastic loss is also ablated qualitatively.
As shown in the visualized mesh in Fig.~\ref{ch:deformtoon3d:fig:ablate_elastic}, the elastic loss is effective in preventing discontinuous deformation and leads to smoother geometry in the styled space.

\input{tabs_and_figs/fig_ablation_elastic_loss}

\input{tabs_and_figs/tab_identity_preservation}

\input{tabs_and_figs/tab_fid}

\subsection{Additional Quantitative and Qualitative Results}
The detailed breakdown of toonification fidelity and quality are shown in Tab.~\ref{tab:id_loss_supp} and Tab.~\ref{tab:fid_supp} respectively.
We include more qualitative experiment results here. In Fig.~\ref{fig:qualitative_comparisons_supp} we include more comparisons with the baseline methods, which demonstrates that \METHODNAME{} produces better quality against existing methods. In Fig.~\ref{fig:additional_inversion_supp} we show more toonification results over real images. The proposed methods yield plausible results with consistent identity preservations.
We further include the stylized texture and shape pair in Fig.~\ref{fig:texture_and_surface_supp} and validate that our method produces high-quality stylization over both texture and shape.

\input{tabs_and_figs/fig_domain_adaptation_supp}
\input{tabs_and_figs/fig_real_image_stylization_supp}
\input{tabs_and_figs/fig_img_and_surface_supp}

%% file: tabs_and_figs/fig_3dmm_animation.tex
% !TEX root = ../main.tex

\begin{figure}[h!]
\begin{center}
  \includegraphics[width=1\linewidth]{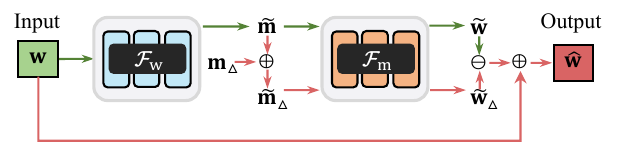}
\end{center}
\vspace{-2mm}
  \caption{Inference pipeline of the proposed animation pipeline.}
\label{fig:3dmm_animation}
\vspace{-2mm}
\end{figure}

%% file: tabs_and_figs/tab_ablation_styles.tex
\begin{table}[h]
% \tiny

\begin{center}
% \vspace{-3mm}
\caption{\textbf{Ablation on the number of styles.}}
\begin{tabular}{l|c|c|c|c}
\hline
\# styles per model   & 1   &  2 & 5 & 10   \\  
\hline
Identity similarity$\uparrow$  &	0.795	& 0.776		&  0.784 & 	0.781 \\ 
\hline
FID$\downarrow$  & 27.5	 & 28.1 & 27.9 &	27.6   \\ 
\hline
\end{tabular}
\label{ch:deformtoon3d:tab:ablate_styles}
\end{center}
% \vspace{-7mm}

\end{table}

%% file: tabs_and_figs/fig_ablation_elastic_loss.tex
\begin{figure}[h!]
\begin{center}
  \includegraphics[width=1\linewidth]{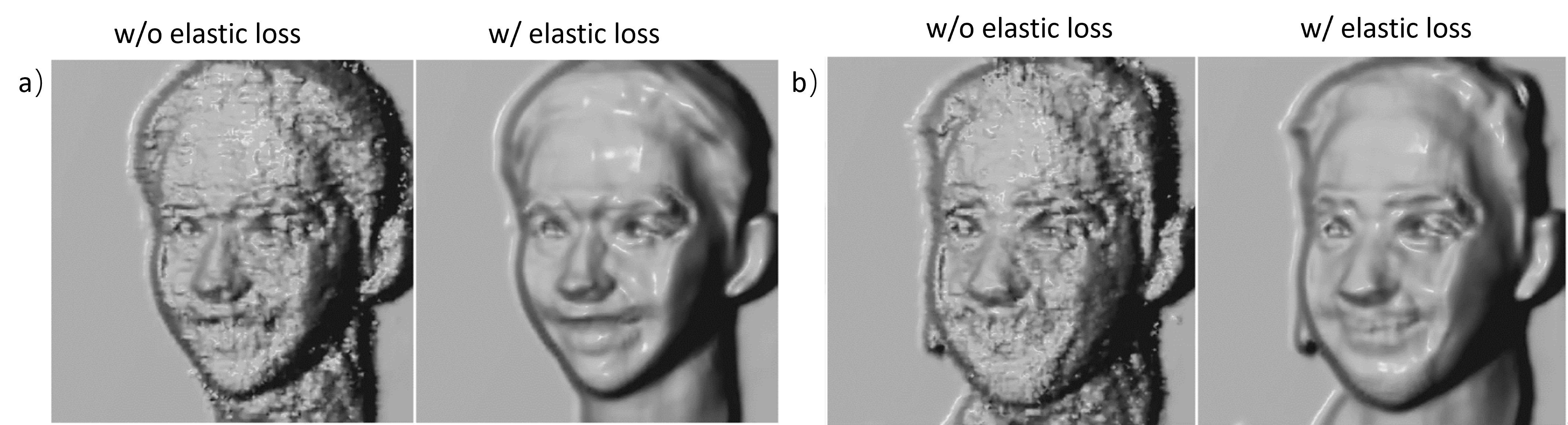}
\end{center}
% \vspace{-2mm}
  \caption{\textbf{Ablation on elastic loss.} Without \textit{v.s.} with elastic loss for a) female and b) male, respectively.}
\label{ch:deformtoon3d:fig:ablate_elastic}
% \vspace{-2mm}
\end{figure}

%% file: tabs_and_figs/tab_identity_preservation.tex
% !TEX root = ../main.tex
\begin{table}
\small
\caption{\textbf{Quantitative evaluation in terms of identity similarity$\uparrow$}. \METHODNAME{} achieves the best identity consistency over all the $10$ styles.}
\label{tab:id_loss_supp}
\vspace{-4mm}
\begin{center}
\begin{tabular}{l|c|c|c|c}
\hline
% Domains   & CIPS-3D  &  E3DGE & StyleGAN-NADA & Ours   \\ 
Domains   & CIPS-3D  &  E3DGE & NADA & Ours   \\ 
\hline
Pixar	& 0.765	& 	0.748	& 0.564	& 		\textbf{0.812}   \\ 
Comic		& 0.643	& 	0.614		& 0.496 & 	\textbf{0.729 }  \\ 
Slam Dunk		& 0.672		& 0.765		& 	0.552	& \textbf{0.780}  \\ 
Caricature I		&  0.648	& 	0.592	& 0.455	& \textbf{0.708}   \\ 
Caricature II		&  0.655 	& 	0.698 	& 0.538	& 		\textbf{0.785}   \\ 
Caricature III		&  0.637 	& 	0.644		& 0.495	& 	\textbf{0.725}   \\ 
Croods		& 0.796		&  0.831		& 0.626
		& \textbf{0.860}   \\ 
Shrek		& 0.708		& 0.794		& 0.599		& \textbf{0.835}   \\ 
Rapunzel		& 0.603		& 0.696		& 0.564	& 	\textbf{0.782}   \\ 
Hiccup		& 0.684		& 0.688	& 	0.464		& \textbf{0.796} 	  \\ 
\hline  
Average		& 0.681	& 	0.707	& 0.535	& 		\textbf{0.781}  \\ 
\hline
\end{tabular}
\end{center}
\vspace{-5mm}
\end{table}

%% file: tabs_and_figs/tab_fid.tex
% !TEX root = ../main.tex
\begin{table}
\small
\caption{\textbf{Quantitative evaluation in terms of $\mathrm{\textbf{FID}}$}$\downarrow$. \METHODNAME{} achieves the best $\mathrm{FID}$ over $9$ of the $10$ styles.}
\label{tab:fid_supp}
\vspace{-4mm}
\begin{center}
\begin{tabular}{l|c|c|c|c}
\hline
Domains   & CIPS-3D  &  E3DGE & NADA & Ours   \\ 
\hline
Pixar	& 33.6 &	36.8 &	39.9 &	\textbf{21.5}   \\ 
Comic	& 61.9 &	44.8 & 70.9	 &	\textbf{33.3}   \\ 
Slam Dunk &	78.1 &	41.8 & 75.9 &	\textbf{37.3}   \\ 
Caricature I &	28.7 &	30.1 & 52.9 &		\textbf{16.0}   \\ 
Caricature II &	76.7 &	58.1 & 102.6 &		\textbf{56.4}   \\ 
Caricature III &	47.1 &	\textbf{25.8} & 54.8 &		27.2   \\ 
Croods &	36.9 &	 30.9	 & 58.5 &	\textbf{22.5}   \\ 
Shrek &	36.0 &	32.1		 & 47.2 & \textbf{20.3}   \\ 
Rapunzel &	65.0 &	30.5	 & 44.1 &	\textbf{17.2}   \\ 
Hiccup &	42.3 &	32.8	 & 46.6 &	\textbf{24.6}   \\ 
\hline   
Average &	50.6 &	34.0	 & 59.3 &	\textbf{27.6}   \\ 
\hline
\end{tabular}
\end{center}
\vspace{-4mm}
\end{table}

%% file: tabs_and_figs/fig_domain_adaptation_supp.tex
% !TEX root = ../main.tex

\begin{figure*}[h!]
% \vspace{-3mm}
\begin{center}
  \includegraphics[width=1\linewidth]{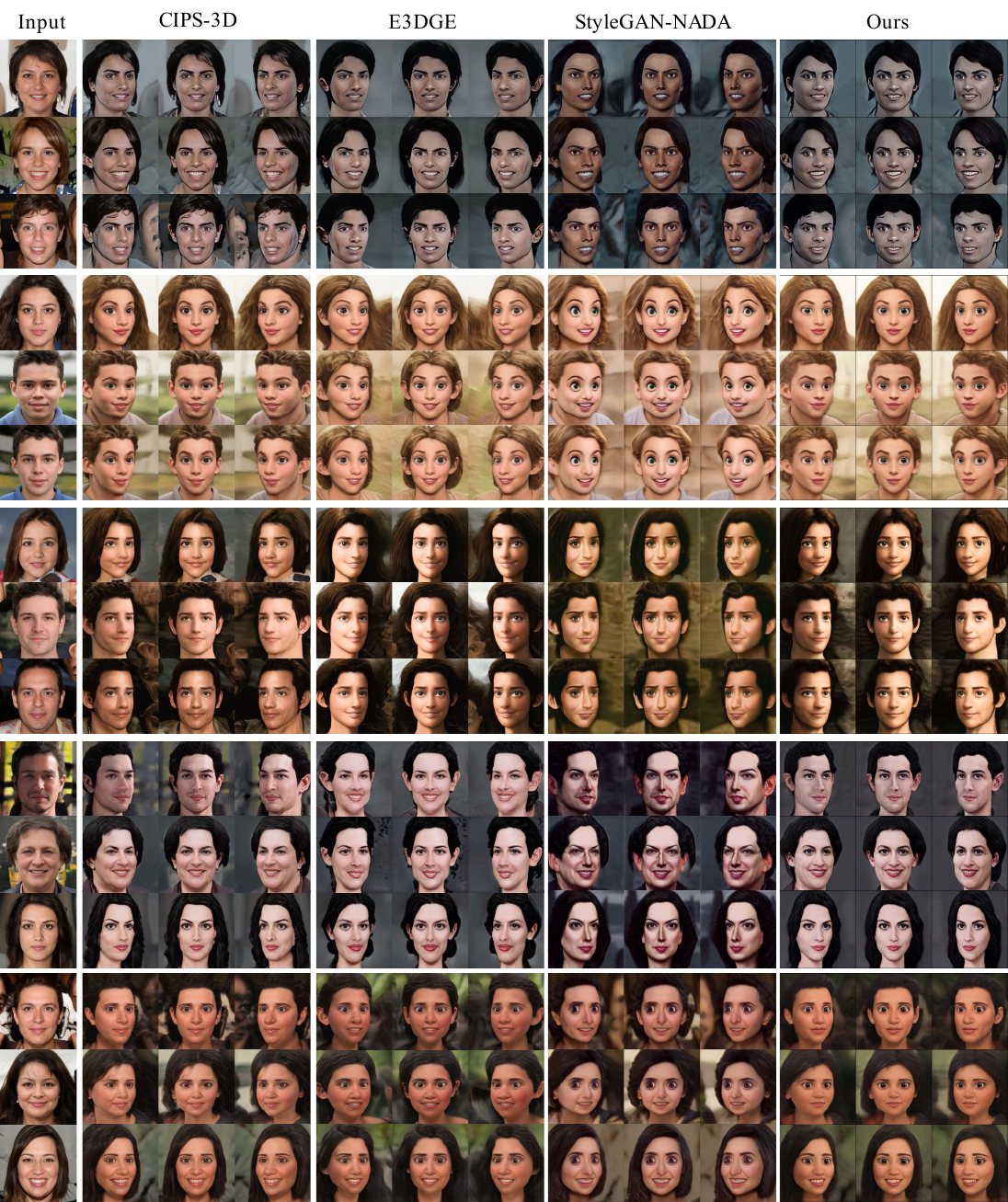}
\end{center}
% \vspace{-2mm}
  \caption{\textbf{Additional qualitative comparisons with baseline methods.} \METHODNAME{} produces better performance against all baselines regarding toonification fidelity, diversity, and identity preservation. Better zoom in.
  }
% \vspace{-2mm}
\label{fig:qualitative_comparisons_supp}
\end{figure*}

%% file: tabs_and_figs/fig_real_image_stylization_supp.tex
% !TEX root = ../main.tex
\begin{figure*}[h!]
% \vspace{-3mm}
\begin{center}
  \includegraphics[width=1\linewidth]{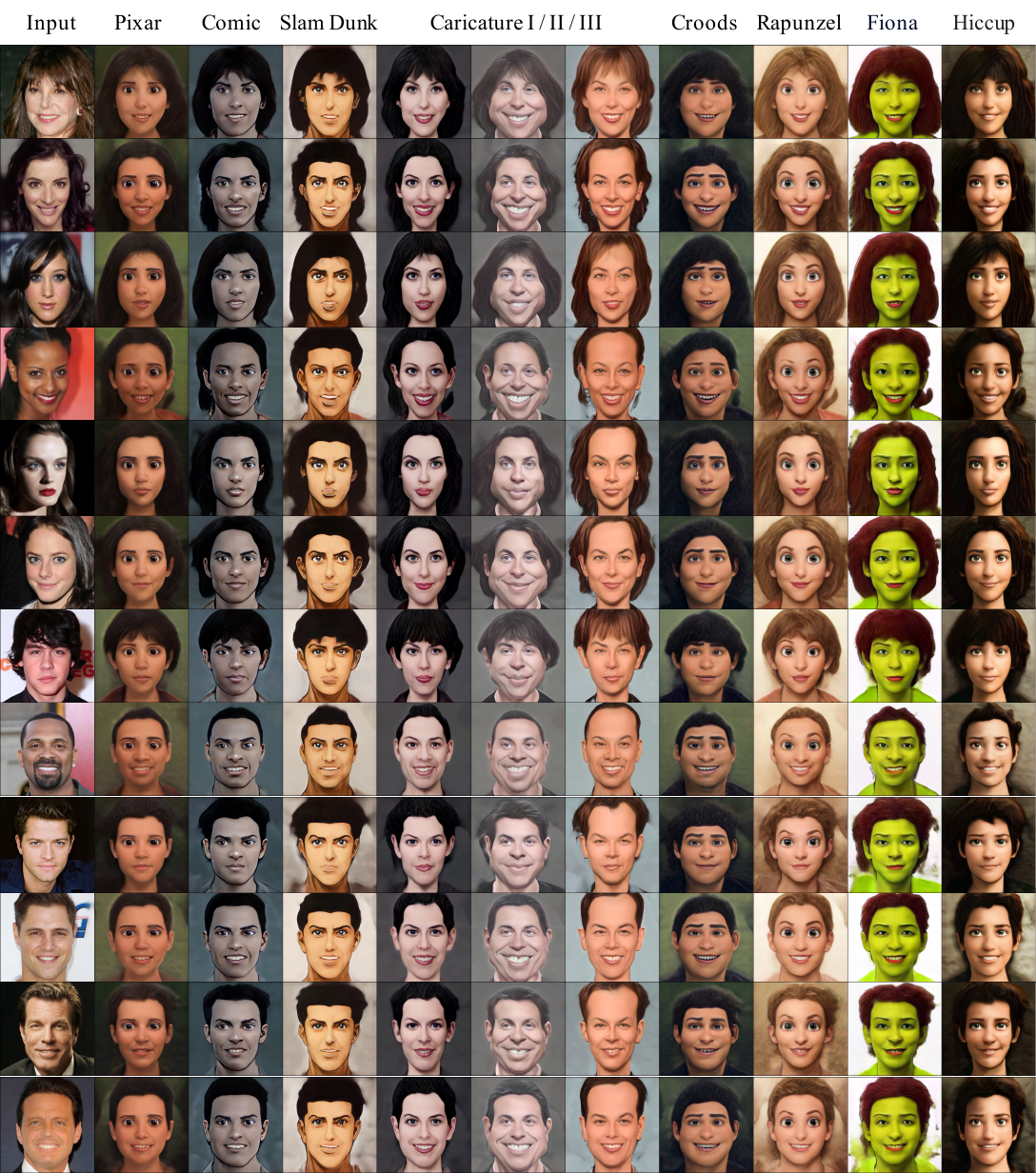}
\end{center}
  % \vspace{-4mm}
  \caption{
 \textbf{Additional results of \METHODNAME{} on the real images.} Our method enables multiple styles toonification with a single model, where both the texture and the geometry matches the target domain.}
  % \vspace{-4mm}
\label{fig:additional_inversion_supp}
\end{figure*}

%% file: tabs_and_figs/fig_img_and_surface_supp.tex
% !TEX root = ../main.tex

\begin{figure*}[h!]
% \vspace{-3mm}
\begin{center}
  \includegraphics[width=1\linewidth]{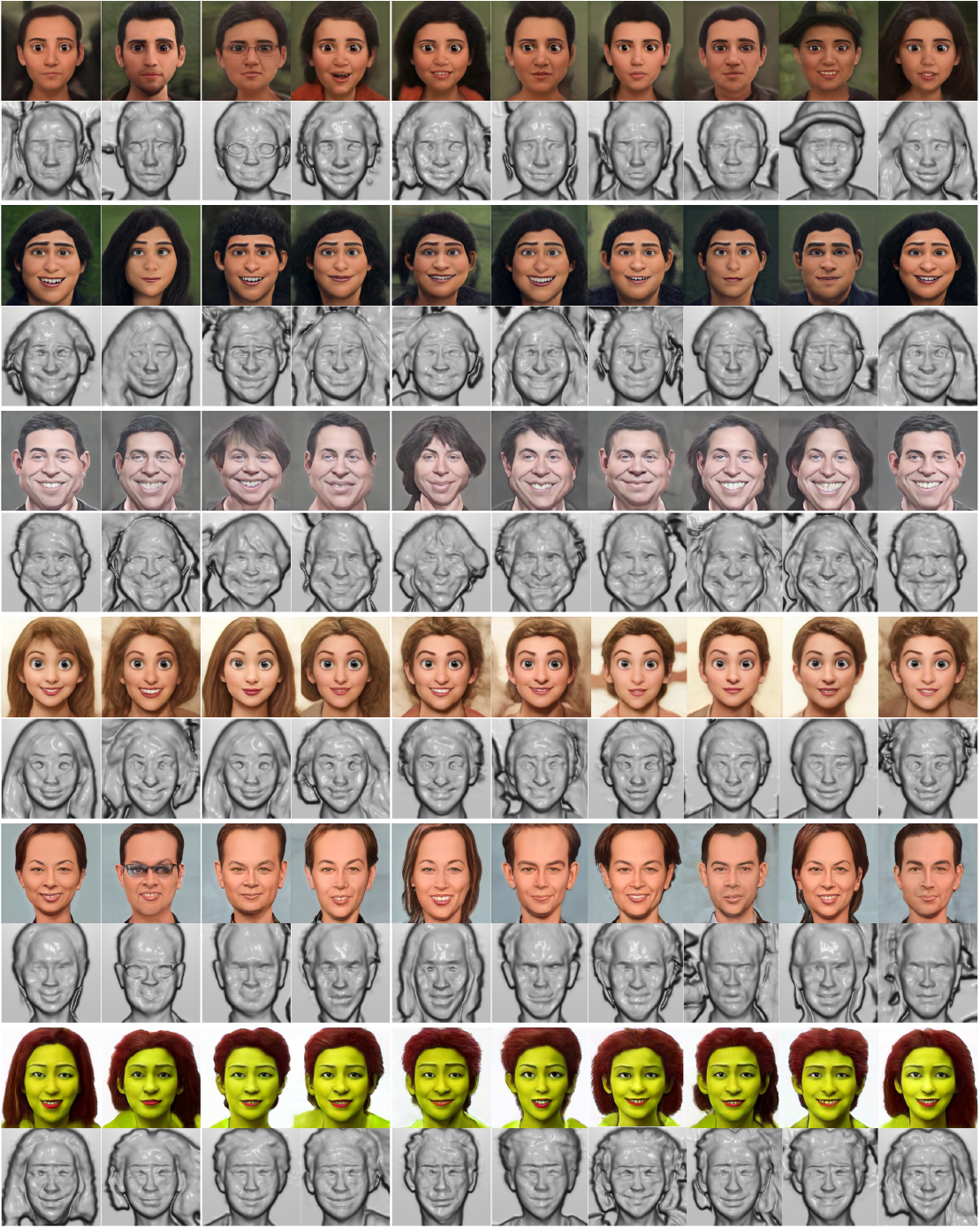}
\end{center}
  % \vspace{-4mm}
  \caption{ 
  \textbf{
  Additional results of \METHODNAME{} with stylized texture and shape.
  }
  }
  % \vspace{-4mm}
\label{fig:texture_and_surface_supp}
\end{figure*}